\documentclass[letterpaper, 10 pt, conference]{ieeeconf}  

\IEEEoverridecommandlockouts                              

\usepackage{cite}
\usepackage{algpseudocode}
\usepackage{algorithm}
\usepackage{graphicx}
\usepackage{cite}
\usepackage{tabularx,booktabs}
\usepackage{amsmath,amsfonts,amssymb}
\usepackage{array}
\usepackage[caption=false,font=normalsize,labelfont=sf,textfont=sf]{subfig}
\usepackage{textcomp}
\usepackage{stfloats}
\usepackage{url}
\usepackage{verbatim}
\usepackage{textcomp}
\usepackage{xcolor}
\usepackage{url}
\usepackage{hyperref}
\usepackage{siunitx}
\usepackage{caption}
\usepackage{subfig}
\usepackage{multirow}
\usepackage{lineno}
\usepackage{xspace}

\title{\LARGE \bf
Performance Comparison of Deep RL Algorithms for Mixed Traffic Cooperative Lane-Changing}

\author{Xue Yao*, Shengren Hou, Serge P. Hoogendoorn, and Simeon C. Calvert 
\thanks{Xue Yao, Serge P. Hoogendoorn, and  Simeon C. Calvert are with the Department of Transport \& Planning, Delft University of Technology, Delft, The Netherlands.  
        Shengren Hou is with the Faculty of Electrical Engineering, Mathematics \& Computer Science, Delft University of Technology, Delft, the Netherlands. 
        {\tt\small x.yao-3@tudelft.nl}}%
}

\begin{document}

\maketitle

\pagestyle{empty}
\thispagestyle{empty}
\begin{abstract}
Lane-changing (LC) is a challenging scenario for connected and automated vehicles (CAVs) because of the complex dynamics and high uncertainty of the traffic environment. This challenge can be handled by deep reinforcement learning (DRL) approaches, leveraging their data-driven and model-free nature. Our previous work proposed a cooperative lane-changing in mixed traffic (CLCMT) mechanism based on TD3 to facilitate an optimal lane-changing strategy. 
This study enhances the current CLCMT mechanism by considering both the uncertainty of the human-driven vehicles (HVs) and the microscopic interactions between HVs and CAVs. The state-of-the-art (SOTA) DRL algorithms including DDPG, TD3, SAC, and PPO are utilized to deal with the formulated MDP with continuous actions. Performance comparison among the four DRL algorithms demonstrates that DDPG, TD3, and PPO algorithms can deal with uncertainty in traffic environments and learn well-performed LC strategies in terms of safety, efficiency, comfort, and ecology. The PPO algorithm outperforms the other three algorithms, regarding a higher reward, fewer exploration mistakes and crashes, and a more comfortable and ecology LC strategy. The improvements promise CLCMT mechanism greater advantages in the LC motion planning of CAVs.  
\end{abstract}

\section{Introduction}
Connected and automated vehicle (CAV) technologies offer a promising solution to improve traffic safety, efficiency and reduce traffic emissions as CAV driving maneuvers can be designed and controlled for certain purposes. According to motion planning of Society of Automotive Engineers (SAE) Level 4 or Level 5 \cite{Yao2021cooperative}, an autonomous vehicle should know when and how to make the proper decision as well as execute the action safely under various traffic scenarios. Nevertheless, defining the robust motion planning strategy is a persistent challenge because of the unpredictable behaviors of surrounding vehicles in mixed traffic (mixed with both CAVs and Human-driven Vehicles (HV)). Specifically, lane-changing (LC) can be a more complex task in motion planning as both longitudinal and lateral behaviors should be considered \cite{shi2019driving, yao2023identification}.

LC motion planning is a sequential decision-making problem with uncertainty. Classical approaches generally formulated LC motion planning as a distributed optimal control problem~\cite{pang2022practical}. The dynamics of autonomous vehicles are formulated by mathematical equations while uncertainty from surrounding vehicles is modeled by a probabilistic model or a set of representative scenarios. Then, model predictive control (MPC) or mixed-integer nonlinear programming (MINLP) is leveraged to find the optimal LC solutions \cite{keskin2020altruistic}. Nevertheless, these approaches usually can not meet the real-time decision requirement because solving MPC or MINLP can be computationally intensive. Moreover, modeling the uncertainty of the traffic environment is challenging. 

By modeling the LC process as a Markov Decision Process (MDP), model-free deep reinforcement learning (DRL) is an alternative approach to overcoming conundrums faced by classical mathematical approaches. DRL approaches aim to learn an optimal policy by trial and error by interacting with the designed environment. After training, DRL approaches can be deployed in real-time, free from the online computation~\cite{sutton2018reinforcement}. For example, deep Q-learning (DQN) was used to learn a vehicle control strategy, in which three-dimensional control actions, namely acceleration, deceleration, and maintaining, were considered \cite{li2020deep, jiang2019interactive}. Jaritz et al. \cite{jaritz2018end} applied an Asynchronous Actor-Critic (A3C) method to learn a vehicle control policy where the control action space includes 32 discrete values. Rather than using discrete action space, Wang et al. \cite{wang2019continuous} formulated the LC problem as an MDP with continuous action spaces and utilized Deep Deterministic Policy Gradient (DDPG) algorithm to define the optimal LC planning strategy. Results demonstrated the effectiveness of continuous MDP action formulation and DRL algorithm solution in addressing LC motion planning in pure CAV traffic by providing optimal strategies with high efficiency. 

LC motion planning in mixed traffic is a more complex task because of the high unpredictability of surrounding HVs \cite{shi2019driving}. For instance, HVs can be non-cooperative in the process of lane changes, such as adopting a hostile motion, which impedes the lane-changing maneuver. The behavior of these HVs is highly stochastic and unpredictable and can not be directly controlled. To solve this challenge, previous research \cite{Yao2021cooperative} proposed a cooperative lane-changing mechanism in mixed traffic (CLCMT) by considering uncertainty in traffic environments. Then, the twin delayed DDPG (TD3) algorithm is used to define the optimal LC motion planning strategy. Results demonstrated the effectiveness of TD3 on the CLCMT problem in terms of safety, efficiency, comfort, etc. Nevertheless, the proposed CLCMT only considers uncontrolled behaviors of HVs, while the interactions among vehicles were ignored, e.g., the collision warnings triggered during the LC process. Besides, there are no silver bullet algorithms since DRL algorithms based on different theories and characteristics would have diverse performances in solving CLCMT tasks. A fair comparison is required to facilitate an optimal option for solving CLCMT problems.

In this paper, we fill the research gaps with two novel contributions: (i) A more realistic CLCMT mechanism is developed, which considers the uncertainty of HVs and the microscopic interactions between HVs and CAVs. (ii) A performance comparison for DRL algorithms including DDPG, TD3, SAC, and PPO is conducted to leverage the capability of different DRL algorithms in solving the formulated CLCMT tasks.

\section{Methodology}
In this section, the cooperative lane-changing problem in mixed traffic is first introduced. Then the DRL-based CLCMT mechanism is presented. 

\subsection{Brief Review of CLCMT}
The process of proposed CLCMT in \cite{Yao2021cooperative} can be outlined in several steps. First, the current speed and desired speed of the target vehicle are acquired. If the former is less than the latter by a threshold of $\epsilon$, a request for lane-changing is triggered. Next, the DRL agent receives traffic states from environment, including leader-follower compositions as well as their gaps in adjacent lanes. These states serve as potential lane-changing scenarios. Then, detailed maneuver control actions behind each potential scenario are executed. This is achieved by DRL-based policies to learn actions including two-dimensional accelerations of $V_{ego}$ (and longitudinal acceleration of $V_{lead}$ and $V_{lag}$ if they are controlled). Based on various lane-changing strategies learned by DRL algorithms, the feedback module computes the utilities of each cooperative lane-changing strategy under different scenarios. According to pre-calculated utilities (including safety, efficiency, comfort, and ecology) and a personalized evaluation function, the feedback module recommends the optimal lane-changing strategy for the decision-making layer where the lane-changing strategy is determined. 


In this CLCMT, vehicles controlled by the DRL agent update their position and speed by following kinematic models, see Equation \ref{eq:kine}, and other vehicles' behavior follows the Intelligent Driver Model (IDM) car-following model \cite{sun2021modeling}, as shown in Equation \ref{eq:CFmodel}.  

\begin{subequations}\label{eq:kine}
\begin{flalign}
  &  x(t_{i+1})=x(t_i)+v_x(t_i)\Delta t + \frac{1}{2} \, a_x \, (\Delta t)^2\\
  &  y(t_{i+1})=y(t_i)+v_y(t_i)\Delta t + \frac{1}{2} \, a_y \, (\Delta t)^2 \\
  &  v_x(t_{i+1})=v_x (t_i)+a_x \, \Delta t \\
  &  v_y(t_{i+1})=v_y (t_i)+a_y \, \Delta t 
\end{flalign}
\end{subequations}
where $x$, $y$ are the positions, $v_x$ and $v_y$ are longitudinal and lateral speed, respectively. $a_x$ and $a_y$ are longitudinal and lateral acceleration, respectively. And $\Delta t = t_{i+1} - t_i$, denotes the time step.

{\small
\begin{subequations}\label{eq:CFmodel}
  \begin{flalign}
  &  a(t_{i+1}) = a_1 \left[1- \left(\frac{v(t_i)}{v_0}\right) ^ \delta \right] - \left(\frac{s^* (v(t_i), \Delta v)}{s_0} \right)^2 \label{con:CFmodel1}\\
  &  s^* \left(v(t_i), \Delta v \right) = s_0 + \max \left( 0, v(t_{i+1})\, T + \frac{v(t_i), \Delta v}{2 \sqrt{a_1\, b_1}} \right) 
  \label{con:CFmodel2}
  \end{flalign}
\end{subequations}}
here $a_1$ is the maximum acceleration/deceleration of the follower, $\delta$ is the acceleration index, $v_0$ is the desired speed and $s_0$ is the minimum distance gap. $s^* \left(v(t_i), \Delta v \right)$ means the desired gap, which is a function of $v(t_i)$ and $\Delta v$ as shown in Equation (\ref{con:CFmodel2}), in which $T$ is the safety time gap and $b_1$ is the comfortable deceleration.

\subsection{Problem Description}
\begin{figure}[!htb]
  \centering
  \includegraphics[width=0.48\textwidth]{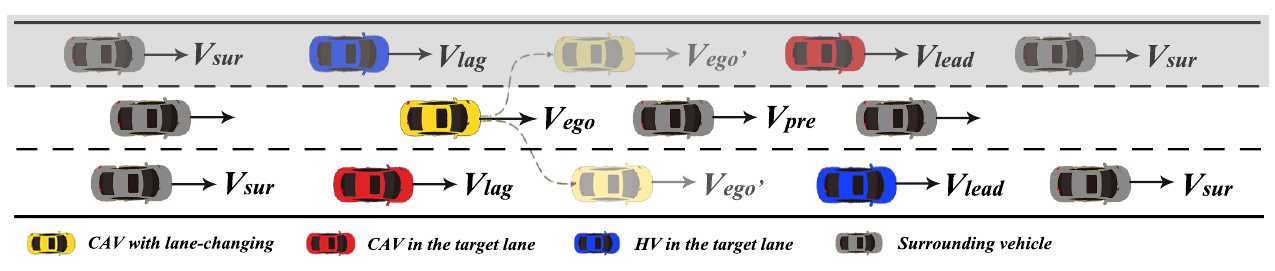}
  \caption{Lane-changing scenario in mixed traffic: An illustrative example}\label{fig:fig1}
\end{figure}

\begin{figure}[htb!]  
  \centering
  \includegraphics[width=0.48\textwidth]{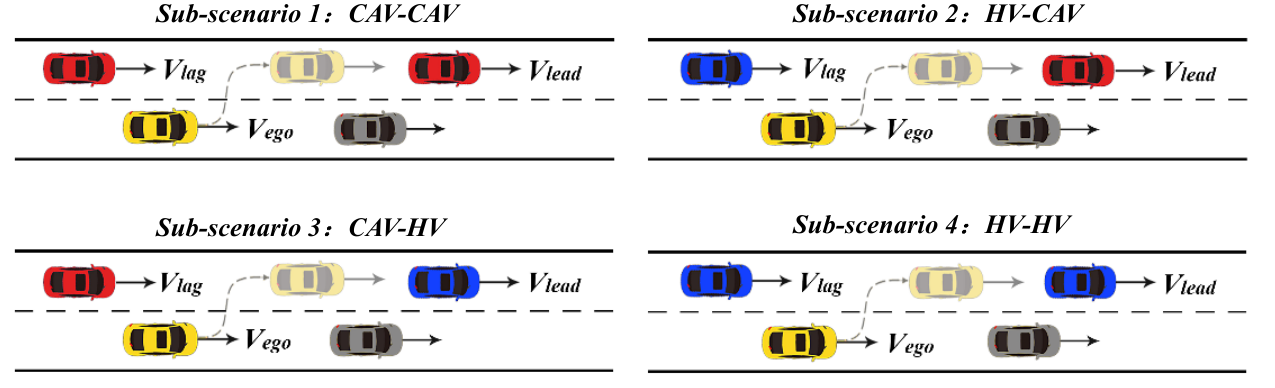}
  \caption{Compositions of leader-follower types.}
  \label{4scenarios}
\end{figure}

Our focus is the pre-calculation procedure in maneuver control of CLCMT where the DRL agent should calculate utilities of all potential lane-changing scenarios and forward results to the feedback module. As shown in Figure \ref{fig:fig1}, the ego vehicle ($V_{ego}$), the leader ($V_{lead}$), and the follower ($V_{lag}$) in the target lane are directly involved in lane-changing and are potential objects in the cooperative control problem. Other vehicles, such as the preceding vehicle in the current lane ($V_{pre}$) and surrounding vehicles in the target lane ($V_{sur}$), are considered in building lane-changing environments. Red and yellow denote CAVs, and blue and gray represent HVs. CAVs can strictly comply with the DRL cooperative maneuver control (CMC) while HVs are provided with cooperative maneuver control recommendation (CMCR) through V2X communication, such as suggested speed and acceleration. Assume that an HV has the probability of $p$ to adopt the CMCR, i.e., behaving as a CAV. The goal of the DRL agent is to learn a feasible LC policy $\pi$ to execute the LC process in a safe, efficient, comfortable, and ecology way. In a two-lane scenario, the leader-follower ($V_{lead}$-$V_{lag}$) that forms a lane-changing gap in the target lane can have four compositions: CAV-CAV, HV-CAV, CAV-HV, and HV-HV, as shown in Figure \ref{4scenarios}. The ego vehicle $V_{ego}$ may face any of them during its DRL learning process. 

\subsection{MDP Formulation}
The CLCMT problem is formulated as a Markov Decision Process (MDP), which is defined by 5-tuple $(S,A,P,R,\lambda)$, where $S$ represents the set of system states, $A$ is the set of action, $P$ and $R$ denote the state transition probability and the reward function, respectively. $\lambda$ is the discount factor used to balance the consequences caused by the uncertainty of future reward. Based on the principles of MDP, the state space, action space, and reward function of the proposed DRL-based CLCMT are provided as follows.

\begin{sloppypar}
(1) \emph{State Space}: the state includes current positions and speeds of the target vehicle $O_1(x_{ego},y_{ego},v_{xego},v_{yego})$, the leader $l_1 (x_{lead},y_{lead},v_{lead})$ and the follower $l_2(x_{lag},y_{lag},v_{lag})$, the surrounding vehicles $s_1 (x_{sur1},y_{sur1},v_{sur1})$ and $s_2 (x_{sur2}, y_{sur2}, v_{sur2})$, respectively. The state space of the four sub-scenarios can be expressed as: 
\end{sloppypar}

\begin{equation}
  s = (O_1, l_1, l_2, s_1, s_2) \in S
\end{equation}

(2) \emph{Action Space}: The action space in this MDP refers to the feasible range of acceleration/deceleration that a controllable vehicle can take. For sub-scenario 1, actions in each lane-changing activity include the acceleration of the target vehicle, the new leader, and the new follower, that is:

\begin{equation}
  a_1 = (a_{egox}, a_{egox},a_{egoy},a_{lead},a_{lag}) \in A_1 \label{con:action1}
\end{equation}

Accordingly, the action space of sub-scenarios 2 - 4 can be shown as follows:

\begin{flalign}
  & a_2=({a_{egox},a_{egoy},a_{lead}}) \in A_2 \\
  & a_3=({a_{egox},a_{egoy},a_{lag}}) \in A_3 \\
  & a_4=({a_{egox},a_{egoy}}) \in A_4 
\end{flalign}

(3) \emph{Reward Function}: Since small deviation in driving behavior may lead to serious consequences, reward functions should be rigorous to ensure robust lane-changing behavior. We consider factors related to safety, comfort, fuel consumption, and emissions in reward function design. 

The safety-related reward is given in Equation \ref{eq15}. As the DRL agent's actions are driven by the pursuit of rewards at each step, the first component of Equation \ref{eq15} is designed to encourage the DRL agent to keep moving forward by offering reasonable rewards. The second component is designed to punish collision. When the agent fails to meet security conditions, i.e., $d_{tar} \leq l_{veh}$, a large negative reward will be given.  $d_{tar}$ represents the current distance between $V_{lead}$ and $V_{ego}$, calculated by the positional difference between two vehicles and a minimum distance $d_0$, see Equation \ref{eq16}. 
\begin{subequations} 
\begin{equation} \label{eq15}
    \mathcal{R}_s =
\begin{cases} 
\alpha \sum^I_{i=1} \Delta x_{ego} + \beta, & \mbox{otherwise} \\
-c,  & d_{tar} \leq l_{veh} 
\end{cases}
\end{equation}
\begin{equation} \label{eq16}
    d_{tar} = x_{lead} - x_{ego} + (v_{lead} - v_{ego})t + \frac{1}{2}(a_{lead} - a_{ego})t^2 + d_0
\end{equation}
\end{subequations}
here, $\Delta x_{ego}$ represents differences of $x_{ego}$ between two timestamps, $l_{veh}$ stands for the length of $V_{ego}$; $t$ is the time at which the lane-change manoeuvre occurs; $\alpha, \beta, \kappa$, and $c$ are coefficients.

Sometimes collisions can be avoided by adjusting the action in time. Thus, we design a collision-check rule to punish risky behavior in the LC process. The judgments involve position and speed, shown as formula \ref{con:warning}. A triggering of the rule is counted as a warning, and the agent gets negative rewards accordingly. 

\begin{subequations} \label{con:warning}
\begin{equation}
\begin{aligned}
& {\rm if} \ \Delta x_i (t) \leq d_0 + \Delta v_i \, \Delta t + \frac{1}{2} \, \Delta a_i \, (\Delta t)^2, \\
& {\rm then} \ a_i (t+1) \leq a_{i+1} (t) - a_s.
\end{aligned}
\end{equation}
\begin{equation}
    R_r = -w
\end{equation}
\end{subequations} 
where $a_i (t+1)$ is the acceleration/deceleration of the follower at $t+1$, and $a_{i+1}(t)$ means the acceleration/deceleration of the leader at $t$; $a_s$ denotes the additional safety room. 

Smooth transitions during lane-changing can provide comfortable experience for CAV users. As such, a comfort reward function $\mathcal{R}_c$ is designed to penalize abrupt jerks and extensive yaws, as shown below. 

\begin{subequations} 
\begin{equation}\label{eq17}
    \varphi = \frac{da}{dt}
\end{equation}
\begin{equation}\label{eq18}
    \theta = \arctan\frac{v_{egoy}(t)}{v_{egox}(t)} - \arctan\frac{v_{egoy}(t-1)}{v_{egox}(t-1)}
\end{equation}
\begin{equation}\label{eq19}
    \mathcal{R}_c = -b_1|\varphi| - b_2|\theta|
\end{equation} 
\end{subequations} 
where $\varphi$ stands for the acceleration/deceleration changing rate of controlled vehicle(s). $\theta$ indicates the yaw changing rate, calculated by the differences between yaws of two adjacent timestamps. $b_1$ and $b_2$ are coefficients.

The reward function of fuel consumption and emissions estimation can be shown as follows. For more details of the model as well as coefficients and default values please refer to \cite{nie2013eco}. $\kappa$ is an adjustment coefficient.

\begin{equation}
  R_f = - \kappa \, T_F
\end{equation}

The positional deviation reward is designed to effectively guide the DRL agent ($V_{ego}$) to promote correct lane-changing directions and be alignment with the centerline of the target lane. The reward function for lateral deviation can be formulated as formula \ref{eq21}. The smaller the lateral deviation, the closer it is to the target centerline. $\varrho, \delta, \zeta$, and $\omega$ are constants, serving as tuning parameters.

\begin{equation} \label{eq21}
    \mathcal{R}_l =
\begin{cases} 
\omega |\Delta d_{lat}|, & \mbox{otherwise} \\
\varrho (|\Delta d_{lat}| - \theta)^2 + \zeta,  & |\Delta d_{lat}| \leq 0.5 
\end{cases}
\end{equation}

The total reward $R$ is the sum of all aforementioned awards, as shown in equation \ref{con:total reward}. All the coefficients in reward are determined via sensitivity analysis in the experiments. 

\begin{linenomath}
  \begin{equation}
  R = R_s + R_r + R_c + R_f + R_l
  \label{con:total reward}
  \end{equation}
\end{linenomath}

\subsection{DRL Policy-based Algorithms}
The formulated MDP consists of continuous state and action spaces, which are difficult to solve by classical RL algorithms, such as $Q$-learning, due to their poor scalability features \cite{sutton2018reinforcement}. By leveraging the generalization and fitting capability of DNNs, DRL algorithms have shown good performance when dealing with this challenge. In valued-based DRL algorithms, the action-state function $Q$ is iteratively updated to indirectly define a deterministic policy, for which the foundation is the Bellman optimality equation:

{\small
  \begin{equation}
  Q^*(s, \, a) = R(s, \, a) + \gamma \, \sum_{s' \in S} P \left(s' \, | \, s, \, a \right) \, \max_{a' \in A} \, Q^* \, \left(s', \, a'\right)
  \end{equation}}

Here, the optimal policy can be derived as $\pi ^* (s) = \arg \max_{a' \in A} Q^*(s, \, a)$ when the optimal value function $Q^*(s, \, a)$ is estimated. Value-based DRL algorithms use DNNs to approximate the $Q$-function, dealing with continuous state spaces. However, continuous action MDP problems require a full scan of the action space when executing policy improvement, i.e.,  $\pi ^* (s) = \arg \max_{a' \in A} Q^*(s, \, a)$, leading to a
dimensionality problem. Instead, policy-based DRL algorithms directly search for the optimal policy, which is usually modeled with a parametric function, denoted as $\pi_{\theta}(s \, |\,a)$.Based on the policy gradient theorem, the policy gradient is expressed as $\nabla_{\theta} J(\theta) = \mathbb{E}_{\pi} \left[Q^{\pi}(s, \, a) \, \nabla_{\theta} \, \ln \pi_{\theta} \, (a\,| \,s) \right] $, $\theta$ can be updated towards the direction suggested by $\nabla_{\theta} J(\theta)$ to find the policy $\pi_{\theta}$ that leads to the highest expected returns. 

We compare the performance of DRL algorithms in solving the CLCMT problem formulated MDP problem, including two off-policy, deterministic algorithms, i.e., DDPG, TD3, and two stochastic algorithms, i.e., Soft Actor-Critic (SAC), Proximal Policy Optimization (PPO). In general, DRL algorithms interact with the environment to collect sequential data, which is then used to update the critic DNNs parameters based on the temporal difference (TD) algorithm. Then, the critic network is used to update actor DNNs parameters based on policy gradient theory. A more detailed explanation of policy-based algorithms can be found in \cite{yu2020policy}.

\section{Experiments and numerical results}
In this section, we first introduce the environment settings and training process of CLCMT. Then experimental results and discussions are provided.

\subsection{Experimental Settings}
In the proposed CLCMT training environment, a 150 \text{m} road with two lanes is constructed. The width of each lane is set to be 3.75 \text{m}. The $y$-axis and $x$-axis denote the longitudinal and lateral motion of driving, respectively. The length of each vehicle is $l_{veh}$ and the spacing (distance between the leader and the follower) is denoted as $L$. The position of each vehicle is depicted by $(x,y)$ coordinate at time $t$. The simulation time step is 0.1 $s$, and the original position of the target vehicle is set at $(60,1.875)$. Spacing $L$ in the original traffic state is set according to common gaps in highway traffic flow, which is 30 \text{m}. Accordingly, the original speed of the traffic flow is set as $15m/s$. Parameters in the IDM model are set according to literature experience \cite{treiber2013traffic} and the experimental traffic conditions, as shown in Table \ref{tab:table1}. The probability $p$ of CHVs is set to be $0.5$.

\begin{table}[htb!]
	\caption{Parameters of car-following model}\label{tab:table1}
	\begin{center}
		\begin{tabular}{|c|c|c|c|}
		\hline
		\textbf{Name} & \textbf{Description} & \textbf{Unit} & \textbf{Value} \\
		\hline
		$a_1$  &  Maximum acceleration &  $m/s^2$  & 3\\\hline
		$v_0$  &  Desired speed  &  $m/s$  & 20 \\\hline
		$s_0$  &  Minimum gap  &  $m$  & 2\\\hline
		$\delta$  & Acceleration index  & -  & 4\\\hline
		$T$   & Safety time gap  & $s$ & 1\\\hline
		$b_1$  & Comfortable deceleration & $m/s^2$ & 1.5 \\\hline

		\end{tabular}
	\end{center}
\end{table}

As for the hyper-parameters of DRL algorithms, the reply buffer size is set as 50000, with the length of 10000 steps as warming up. The discount factor of future rewards and the learning rate are set as 0.995 and $6 \times 10^5$, respectively. The network width is 256. The batch size (the number of transitions sampled from reply buffer) is 2000. The exploration noise and smoothing regularization noise (only for TD3) are both set to be 0.5, aiming to improve robustness. The training process is set to be 5000 episodes to guarantee convergence. To minimize the impact of perceptual information errors and improve learning strategies' robustness, we add some noise to the original state when resetting the environment. The reset noises are uniformly distributed within the range of [0,1],[0,0.5], and [0,2] for $x$, $y$, and $v$, respectively. Proper parameters for the designed reward functions are defined by empirical theories and fine-tuning in the pre-training process.

\subsection{Results and Discussions}
\subsubsection{Performance of DRL-based algorithms}
\begin{figure*}[htb!]
  \centering
  \includegraphics[width=0.8\textwidth]{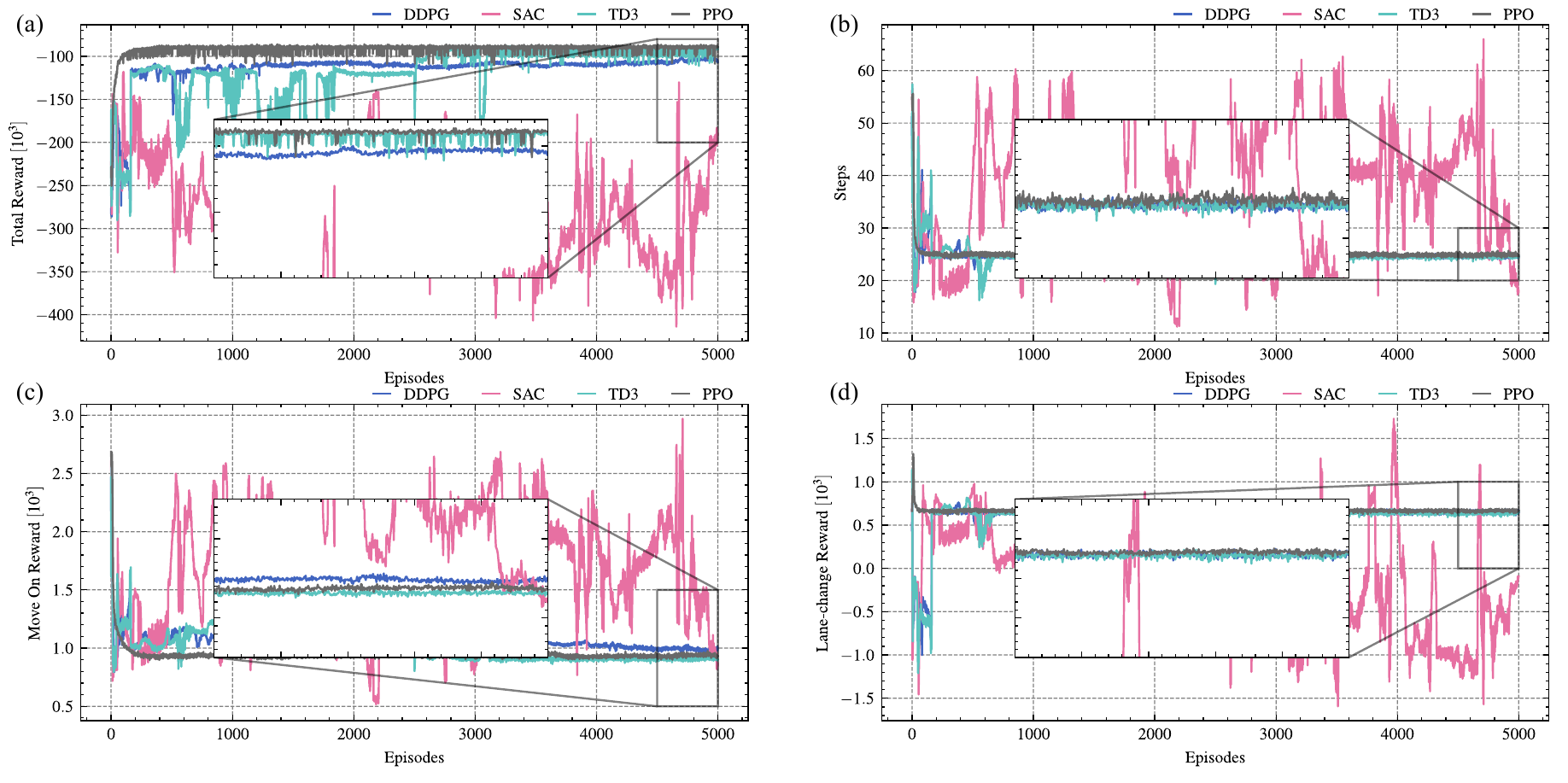}
  \caption{Average (a) total reward, (b) time steps, (c) move-on reward, (d) lane-changing reward}
  \label{fig:cav-cav-1}
\end{figure*}

Based on the settings mentioned above, the agent was trained in simulated scenarios under designed rewards. The total reward shows the robustness of algorithms in solving this CLCMT problem. 
Figure \ref{fig:cav-cav-1} illustrates the average total reward, time steps of the LC process, move-on reward, and lane-changing reward of the four DRL algorithms during the training process. For the PPO algorithm, the average total reward increased rapidly in the first 200 episodes, see the gray line in Figure \ref{fig:cav-cav-1}a. Then it converged with small fluctuations and finally maintained the highest total reward among the four DRL algorithms. For DDPG and TD3 algorithms, the reward substantially decreased between 50 and 200 episodes, then increased rapidly to a high level, which converged after 600 episodes and 3200 episodes, respectively. DDPG has a more stable trend than TD3 during the training process. This is because compared to DDPG, TD3 introduces an extra action noise for the trained policy, which can avoid local optimality but impact the training stability. To this end, TD3 showed a higher total reward than DDPG after converging, see the dark blue and blue lines in Figure \ref{fig:cav-cav-1}a. The constantly fluctuating pink lines in Figure \ref{fig:cav-cav-1}a-d show that the SAC can not converge during the training process, indicating a failure to deal with the CLCMT problem. More specifically, all three algorithms, i.e., PPO, DDPG, and TD3, converged at around 25 timesteps, meaning that they used 2-3s to complete this LC maneuver, see Figure \ref{fig:cav-cav-1}b. These successfully trained algorithms also have similar lane-changing rewards after convergence, see Figure \ref{fig:cav-cav-1}b, and d, and DDPG obtained higher move-on rewards compared to PPO and TD3 (Figure \ref{fig:cav-cav-1}c).

\begin{figure*}[htb!]
  \centering
  \includegraphics[width=0.8\textwidth]{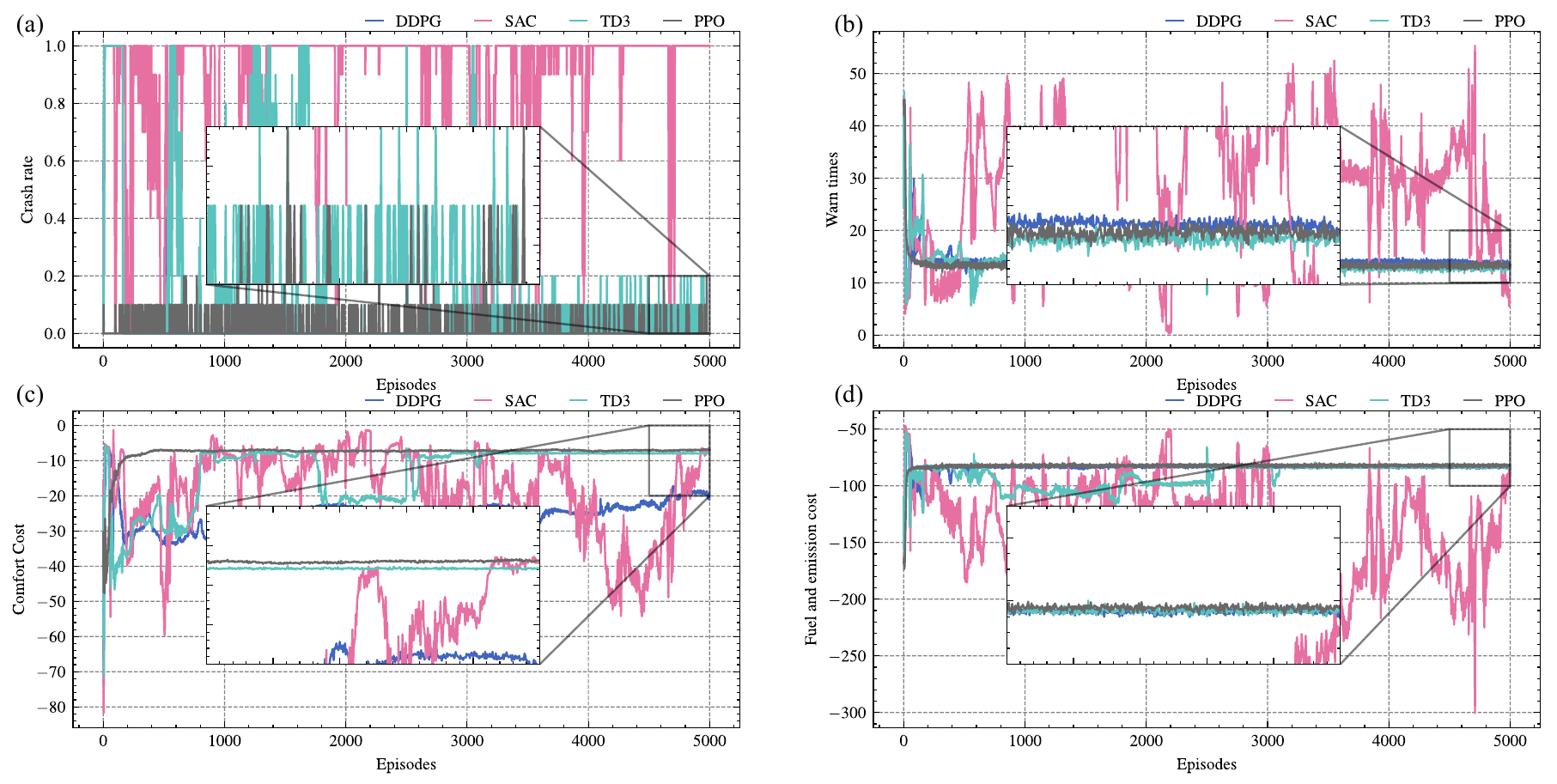}
  \caption{Average (a) crash rate, (b) warning times, (c) comfort cost, (d) fuel consumption and emissions cost}
  \label{fig:cav-cav-2}
\end{figure*}

Figure \ref{fig:cav-cav-2} shows the results of average crash rate, warning times, comfort, fuel consumption and emissions rewards during the training process. In Figure \ref{fig:cav-cav-2}a, the crash rates of PPO, DDPG, and TD3 decreased significantly during the training and can avoid any crashes finally, while SAC did not learn a policy that can prevent crashes. Additionally, DDPG triggered more collision warnings than PPO and TD3 during the training, see Figure \ref{fig:cav-cav-2}b. This is because the DDPG-based agent tends to explore actions located in boundaries compared to PPO and TD3, leading to radical accelerations and thus exhibiting more risky behaviors. Figure \ref{fig:cav-cav-2}c-d reveals that PPO learned the most comfortable and environmentally friendly LC motion strategy compared to other algorithms. 

\begin{table}[!htb]
    \centering
    \caption{Parameters of car-following model}
    \label{tab:utility}
    \begin{tabular}{|c|c|c|c|c|}
        \hline
	\textbf{DRL algorithm} & \textbf{$U_t$} & \textbf{$U_s$} & \textbf{$U_c$} & \textbf{$U_e$}\\
	\hline
	DDPG  & 0.993     & 0.974      & 0          & 0.98 \\\hline
	SAC   & 0          & 0          & 0.392      & 0 \\\hline
	TD3   & \textbf{1} & 0.866      & 0.668      & 0.85 \\\hline
	PPO   & 0.968      & \textbf{1} & \textbf{1} & \textbf{1} \\\hline
    \end{tabular}
\end{table}

To evaluate LC strategy learned by the four DRL algorithms, we adopt four utilities, i.e., the efficiency$(U_t)$, safety level $(U_s)$, comfort level $U_c$, and ecology $(U_e)$ of lane-changing. They are evaluated by completion time of lane-changing, crash rate, comfort reward $R_c$, and fuel consumption and emissions $R_f$, respectively. Table \ref{tab:utility} shows the utility results of each algorithm after convergence (i.e., the last 1000 episodes). Each utility is normalized by the highest and lowest value within that utility. 
Notably, PPO has the highest utilities of safety, comfort, and ecology, and the utility of efficiency is very close to TD3 which performs best in terms of efficiency. DDPG performs well in efficiency, safety, and ecology, however, it has the lowest utility in terms of comfort. TD3 algorithm has outstanding utility in efficiency, with moderate performance in other utilities. Similar to aforementioned analysis, SAC learned the worst strategy in terms of all utilities. Overall, performance comparison of the four algorithms shows that PPO has the best performance when compared with DDPG, SAC, and TD3 algorithms. This is because PPO is an on-policy algorithm, which is more stable than the three off-policy algorithms. Moreover, PPO is theoretically based on Trust Region Policy Optimization (TRPO), which guarantees the parameter updating for each step will increase its performance in the current environment.

\begin{figure}[!ht]
  \centering
  \includegraphics[width=0.4\textwidth]{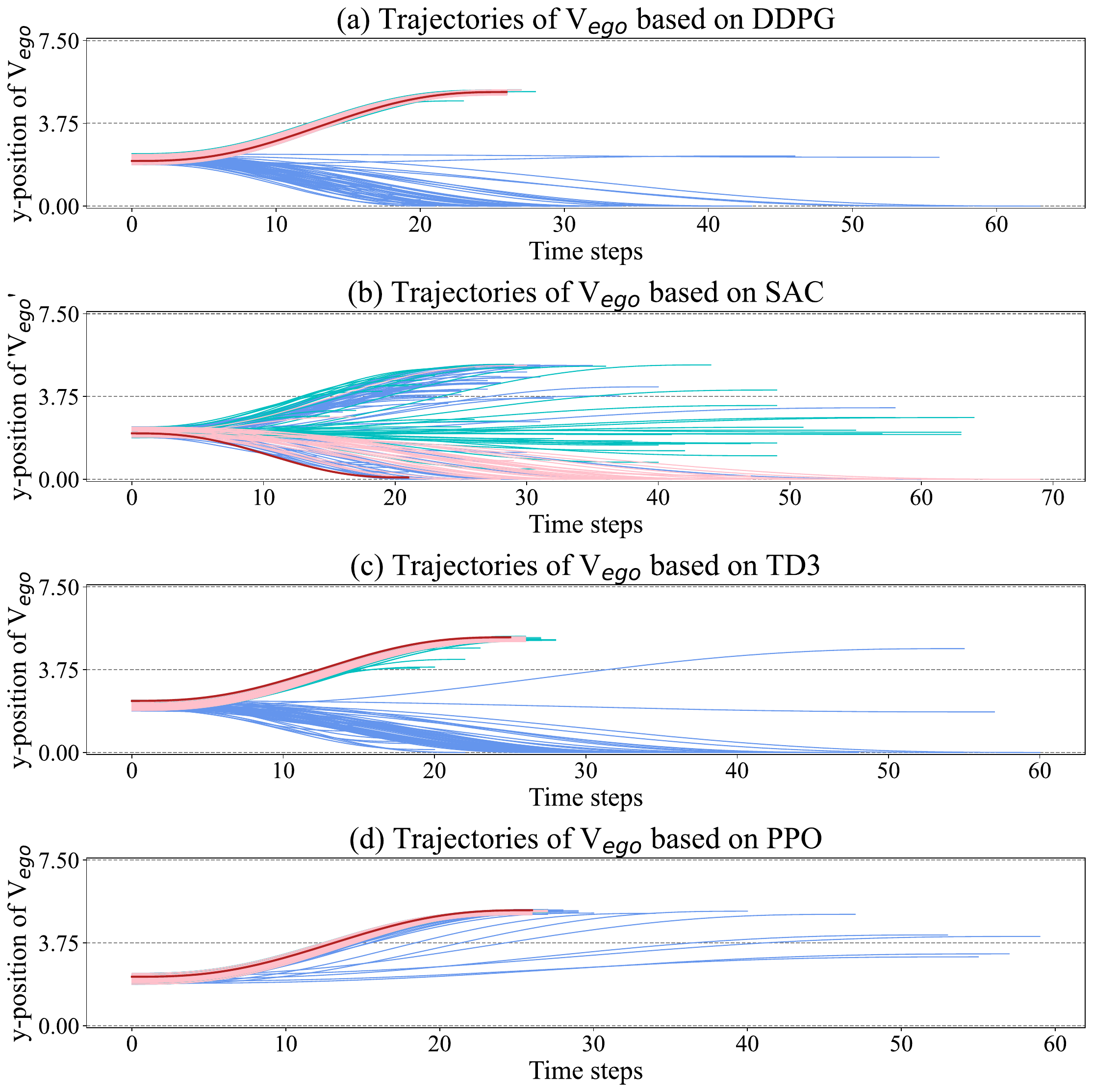}
  \caption{Trajectories of ego vehicle based on (a) DDPG, (b) SAC, (c) TD3 and (d) PPO algorithm}
  \label{fig:traj}
\end{figure}

A series of checkpoints during the training process are selected to further assess trajectories learned by DRL algorithms. As shown in Figure \ref{fig:traj}, the blue, green, and pink lines represent LC trajectories at the first 200 episodes, the intermediate stage, and the last 200 episodes, respectively. Note that the DRL agent explored many trajectories in the wrong direction or failed to complete LC during a certain road section in the first 200 episodes, see blue lines. In the intermediate stage, the DRL agent found the right direction and tried to complete lane-changing, in which trajectories from PPO and DDPG are closer to ideal lane-changing compared to TD3 and SAC. Lane-changing trajectories are further optimized in the last 200 episodes according to designed rewards, and the final optimal lane-changing trajectories from the four algorithms are illustrated by bold red lines. These well-performed trajectories confirm the good capabilities of DDPG, TD3, and PPO algorithms in solving LC motion planning.

\section{Conclusion}
In this study, we first proposed the cooperative lane-changing in mixed traffic (CLCMT) mechanism by considering both the uncertainty of HVs and the microscopic interactions between HVs and CAVs. Then we formulated CLCMT as a Markov Decision Process (MDP) and then conduct a fair performance comparison of four SOTA DRL algorithms for solving CLCMT problems. Experimental results reveal good capabilities of DDPG, TD3, and PPO algorithms for LC motion planning with significantly low crash rates. Specifically, the PPO algorithm outperforms DDPG and TD3 algorithms by providing a safer, more comfortable, and more environmentally friendly lane-changing strategy. Future research may involve heterogeneity of driving behavior in the CLCMT framework, which can help illustrate HV uncertainty more accurately and effectively. 



\bibliographystyle{IEEEtran}
\bibliography{reference}

\end{document}